\documentclass{article}

\usepackage[final]{nips}

\usepackage[utf8]{inputenc} 
\usepackage[T1]{fontenc}    
\usepackage{hyperref}       
\usepackage{url}            
\usepackage{booktabs}       
\usepackage{amsfonts}       
\usepackage{nicefrac}       
\usepackage{microtype}      
\usepackage{amsmath}
\usepackage{booktabs}
\usepackage[colorinlistoftodos]{todonotes}
\usepackage{pifont}
\usepackage{xfrac}
\usepackage{color,soul}

\newcommand{\etal}{\textit{et al.}}

\newcommand\Tstrut{\rule{0pt}{2.6ex}} 

\title{Adaptive additive classification-based loss for deep metric learning}

\author{
	Istvan Fehervari \\
	Amazon\\
	\texttt{istvanfe@amazon.com} \\
	\And
	Ives Macedo \\
	Amazon\\
	\texttt{macedoim@amazon.com} \\
}

\begin{document}

\maketitle

\begin{abstract}
Recent works have shown that deep metric learning algorithms can benefit from
weak supervision from another input modality. This additional modality can be
incorporated directly into the popular triplet-based loss function as distances.
Also recently, classification loss and proxy-based metric learning have been
observed to lead to faster convergence as well as better retrieval results,
all the while without requiring complex and costly sampling strategies.
In this paper we propose an extension to the existing adaptive margin for
classification-based deep metric learning. Our extension introduces a separate
margin for each negative proxy per sample. These margins are computed during
training from precomputed distances of the classes in the other modality.
Our results set a new state-of-the-art on both on the Amazon fashion retrieval
dataset as well as on the public DeepFashion dataset. This was observed with
both fastText- and BERT-based embeddings for the additional textual modality.
Our results were achieved with faster convergence and lower code complexity than
the prior state-of-the-art.
\end{abstract}

\section{Introduction}

Learning representative image embeddings has been the core problem for many
computer vision applications like image search and retrieval, recommendations,
image near-duplicate detection, face or logo recognition to name a few.
Following the advances of deep learning, arguably the most popular technique to
learn image representations is deep metric learning (DML). The idea is to learn
a mapping function that projects images to a lower dimensional latent space
where semantically similar instances are clustered together. Instead of
learning discriminative characteristics of classes in the training data, DML
models learn a function that enables a measure of similarity via distances
(e.g. Euclidean or cosine) allowing them to generalize to unseen classes as
well.

There have been several proposals on how to learn such embedding functions. The
classic ones are based on losses that are computed on pairwise distances. For
example, the contrastive loss~\cite{contrastiveloss} uses matching and
non-matching image pairs, while the triplet-loss~\cite{tripletloss} operates
on a tuple of two instances from the same class (anchor $a$ and positive $p$)
and a third one from a different class (negative $n$), to exploit pairwise
distance relationships. The fundamental problem of these approaches is that,
ideally, one would have to sample every possible pair or triplet during
training, which is computationally intractable. Instead, several sampling
strategies were proposed for triplet-based losses~\cite{schroff2015facenet,
Harwood2017SmartMF, wu2017sampling, wang2019multi, wang2020xbm} to ensure that
only \emph{informative} tuples are being used in the training batch. In fact,
if the sample is too easy, meaning $d_{ap} \ll d_{an}$ where $d_{xy}$ defines
the distance between sample $x$ and $y$, the loss converges to 0, slowing down
training. Too hard samples on the other hand, where $d_{ap} \gg d_{an}$, could
destabilize training and cause the embeddings to collapse into a single point.
Though such sampling strategies help in training DML models, they often require
complex implementations that are hard to parallelize or to compute ahead of
each batch.

In 2017 Movshovitz-Attias \etal~\cite{movshovitz2017no} introduced the idea of
proxies that are learnable class centers substituting the positive and the
negative instances in the loss function during training. Such approach promises
faster converges and requires no sampling. A very similar idea was also
proposed by Snell \etal~\cite{snell2017prototypical} where the class centers
were derived from a larger batch instead. More recently a combination of the
softmax cross-entropy loss and the learnable proxies was shown to achieve even
better results~\cite{normproxies}, with low implementation complexity, quick
convergence, and easily parallelizable operation.

Even though several novel approaches on metric learning losses are published
every year, there is a new line of research suggesting that, with proper
evaluation protocol and hyper-parameter tuning in place, the actual differences
between the performances of these losses are
negligible~\cite{Fehrvri2019UnbiasedEO,
Roth2020RevisitingTS, Musgrave2020AML}. We hypothesize that a potential way to
improve metric learning is by utilizing extra information existing in other
modalities during training. A promising approach was presented last year, when
Zhao et al.~\cite{Zhao2019AWS} used product titles as weak supervision
to establish adaptive margins in the triplet loss function. Although this method
achieves better retrieval performance compared to the triplet loss baseline and
comparable results to a state of the art DML method on a very large Amazon 
fashion dataset, it suffers from slow convergence and requires complex
unparalellizable sampling code. It is also unclear whether this method can be
applied to state-of-the-art DML losses and whether it would achieve state of
the art results on a smaller public dataset.

In this paper, we show how the weakly-supervised adaptive margins can be
applied to cross-entropy-based losses by proposing an adaptive additive angular
margin term in the softmax function. Our method is easy to implement, requires
no sampling, and achieves better performance on the Amazon fashion retrieval benchmark
 dataset and state-of-the-art results on the public DeepFashion dataset, even with lower
embedding dimensions.

The main contributions of this paper are the following:
\begin{itemize}
	\item We empirically demonstrate that an additive angular margin in the
		softmax loss is effective for proxy-based deep metric learning;
	\item We show the advantages of using non-constant margins on the
		negative classes;
	\item We demonstrate that the negative margins can be derived from
		representations of another modality, similarly as
		in~\cite{Zhao2019AWS}, and this improves retrieval performance;
	\item We evaluate the proposed approach on the Amazon fashion retrieval
		dataset as well as on the public DeepFashion dataset and set
		new state-of-the-art results on both.
\end{itemize}

\section{Adaptive Additive Softmax}

Deep metric learning aims to learn an embedding space where the similarity
between the input samples are preserved as distances in the latent space. For
example, metric learning losses such as Contrastive
Loss~\cite{contrastiveloss}, or Triplet-Loss~\cite{tripletloss} are designed to
minimize the intra-class distances and maximize the inter-class distances. More
recent approaches however consider the relationship of all the samples in the
training batch to maximize efficiency~\cite{wang2019multi}. In fact, the key
problem in DML is how to sample informative training samples that yield
near-optimal convergence. Semi-hard mining proposed by Schroff
\etal~\cite{schroff2015facenet} has been widely adopted for many tasks due to
its ability to mine samples online. More recent approaches propose sampling by
weighting distances~\cite{wu2017sampling} or dynamically building class
hierarchies at training time~\cite{hierarchical_triplet}.

Softmax-based losses have been largely applied in face verification tasks,
achieving state of the art results~\cite{Wang2018CosFaceLM, wen_2016,
Liu2017SphereFaceDH}. The advantage of these losses is the decreased emphasis
on the sampling technique at the cost of additional hyper-parameters and
potentially worse generalizations. A theoretical connection between
classification and deep metric learning has been investigated
in~\cite{movshovitz2017no}, where the authors propose to use learnable class
centers during training, which they call \emph{proxies}. Let us consider
the temperature-scaled normalized softmax loss function for proxy-based
metric learning originally defined in~\cite{normproxies}:
\begin{equation}
L_{nt} = - \log
\left (
	\frac{ \exp{ (x^{T} p_{\text{y}} / \sigma ) }  }
	     { \sum_{\text{z} \in Z} \exp{ (x^{T} p_{\text{z}} / \sigma ) } }
\right )
\end{equation}
where $x$ is an L2-normalized embedding corresponding to the output of the last
linear layer of the model. $\text{y}$ is the class label of $x$ of all possible
classes $Z$, and $p_{\text{y}}$ is its respective proxy embedding. The
temperature parameter $\sigma$ is used to scale the logits to emphasize the
difference between classes, thus boosting the
gradients~\cite{Liu2017SphereFaceDH, Wang2018CosFaceLM}.

Similarly to works in the face classification domain~\cite{Wang2018CosFaceLM,
wang2018additive}, an additive large margin term $m$ can be introduced on
the positive pairs in the proxy-based softmax loss to improve class separation
in the latent space, leading to the Large-Margin Cosine Loss (LMCL).

\begin{equation}
L_{\text{LMCL}} = - \log
\left (
	\frac{ \exp{ ((x^{T} p_{\text{y}} - m) / \sigma ) } }
	     {
		\exp{ ((x^{T} p_{\text{y}} - m) / \sigma ) }
		      + \sum_{\text{z} \in Z^{\text{y}}}
				\exp{ (x^{T} p_{\text{z}} / \sigma ) }
	     }
\right )
\end{equation}

where $Z^{\text{y}}$ indicates the set of all classes except $\text{y}$.
As presented later in our experiments, a constant margin already improves the
retrieval results.

However, we can also introduce additive margins on the negative pairs based on
semantic class similarity in another modality, as demonstrated for the
triplet-loss by Zhao \etal~\cite{Zhao2019AWS}. Our proposed loss with the
adaptive additive large margin on the proxy-based classification loss
is the following:

\begin{equation}
L_{\text{Adaptive\_margin}} = - \log
\left (
	\frac{ \exp{ ((x^{T} p_{\text{y}} - m) / \sigma ) } }
	     {
		\exp{ ((x^{T} p_{\text{y}} - m) / \sigma ) }
		+ \sum_{\text{z} \in Z^{\text{y}}}
			\exp{ ( ( x^{T} p_{\text{z}}
				  + (1-x^{T} p_{\text{z}} )d_{\text{yz}} )
				/ \sigma ) }
	     }
\right )
\end{equation}
where $d_{\text{yz}}$ is the Euclidean or cosine distance of the class
representations of $y$ and $z$ in the other modality normalized between 0 and
1.

There are multiple ways of obtaining such class representations, for
example natural language descriptions or sets of attributes of each class can
be used. For the former, one can apply state-of-the-art pre-trained models
like BERT~\cite{Devlin2019BERTPO}. In fact, in this work we use a model that
consists of BERT further trained on Amazon's textual datasets such as product
titles, description, bullet-points, and product reviews, we call this model
AmaBERT. For the attributes, an average vector of all word embeddings of the
attributes could be used. When no such data exists, a pre-trained image
captioning model could be used as a form of knowledge distillation.

The extensions we propose introduce negligible extra computational and memory
costs and, with the class distances pre-computed, one can enable much faster
training when compared to sampling-based approaches.
In subsection~\ref{results}, we present and discuss concrete computational
results and overview performance differences of different approaches
on two benchmark datasets.

\section{Experiments}

We evaluate the impact of our proposed extensions by comparing multiple
variants trained and benchmarked against the same
datasets~\cite{liuLQWTcvpr16DeepFashion,Zhao2019AWS}. Our experiments cover
classification-loss-based variants with different temperatures,
no-/constant-/adaptive-margin~\cite{normproxies,Wang2018CosFaceLM,
wang2018additive}, and the adaptive-margin-based variant that introduces new
modalities~\cite{Zhao2019AWS}. Moreover, we evaluate different feature
extractors for both the image embedding backbone (RestNet50, EfficientNet-B0
and -B1) and the embeddings of textual modality
(fastText~\cite{bojanowski2016enriching}, and the BERT-based
AmaBERT). Models trained with softmax loss have been observed to
produce sparse embeddings, hence with the introduction of a parameterless
layer normalization one can easily threshold the embedding vectors without much
loss on accuracy. In our experiments we also investigate how much the
performance for each of these models suffer when such thresholding is applied.

\subsection{Data}

We build our experimental evaluation on two datasets built from different
sources and use to benchmark previous work~\cite{Zhao2019AWS}.
Both of these datasets were set up to assess image retrieval performance on
product imagery from real world retailers, however they differ in size
and content specifics which we summarize below.

\paragraph*{Amazon Fashion Retrieval dataset.}
The Amazon fashion retrieval benchmark dataset is consisting of 82,465 images sourced from
22,200 fashion products in the Amazon catalog. This dataset was built with
the help of three fashion specialists to remove irrelevant imagery from a much
larger initial collection consisting of 164K products of 84 different types
totalling over 1.4 million images. Details of the steps taken in cleaning the
original collection can be found in~\cite{Zhao2019AWS}. A key characteristic of
this dataset is that it contains associated textual information in addition to
the imagery, sourced from the Product Detail Page. We use the product title as
the additional (textual) modality augmenting each product's images through text
embeddings computed using pre-trained fastText~\cite{bojanowski2016enriching}
and AmaBERT.

\paragraph*{DeepFashion's In-Shop Clothes Retrieval dataset.}
In addition to the larger Amazon dataset, we use the publicly-available
DeepFashion benchmark~\cite{liuLQWTcvpr16DeepFashion} (sourced from
\emph{Forever 21}'s product catalog) to evaluate the performance of our models
in cross-domain retrieval. Specifically, we leverage its
\emph{In-Shop Clothes Retrieval} version for its inclusion of textual attributes
describing product characteristics in addition to imagery spanning 11,735
products and totaling 54,642 images. Mirroring the methodology used
in~\cite{Zhao2019AWS}, our evaluation follows the approach detailed
in~\cite{liuLQWTcvpr16DeepFashion} and Top-K accuracy is reported as the main
evaluation metric.

It is worth noting that, in both of these datasets, classes are associated with
the products themselves, thus comprising of tens of thousands of classes in
stark contrast to the popular CUBS~\cite{WahCUB_200_2011} and
Cars~\cite{cars196} benchmarks with 100 and 98 classes in their test sets,
respectively.

\subsection{Implementation}

All experiments were conducted with a single Tesla V100 GPU in the PyTorch deep
learning framework version 1.4. On the Amazon Fashion Retrieval Dataset we used
SGD with batch size of 75, momentum of 0.9, and learning rate of 0.01 with
exponential decay for 500,000 iterations. We used linear learning rate warmup
for the first 3000 iterations. For feature extractors we used ImageNet
pretrained ResNet50 and EfficientNet-B0 models with embedding size of 2048 and
1280 respectively. For all models, we used the same image input size of
$224\times224$. The fully-connected embedding layers with non-parametric
layer normalization were initialized randomly and were added right after the
last pooling layer in each architecture. The output of the model was L2
normalized. Except where explicitly noted, we used a temperature scaling parameter $\sigma$
of $20$ in the softmax-crossentropy loss. We set $m=0.4$ for all experiments
which included margin that we derived via hyper-parameter tuning.
During training we used class-balanced sampling with $k=5$ images per class.
When training on the In-Shop dataset we used the same setup as
in~\cite{normproxies}.

\subsection{Results and Discussion\label{results}}

\paragraph*{Amazon Fashion Retrieval dataset.} For our experiments on the Amazon
fashion dataset, we set the triplet-based loss with adaptive margin and with
embedding size of 128 as our baseline, as proposed in the respective paper.
However, while reproducing the results presented in~\cite{Zhao2019AWS} using the 
original codebase we obtained slightly better results which we are using here instead.
We found that the original
normalized softmax loss with a 2048-dimensional embedding easily outperforms
this, lifting the Recall@1 from 88.46\% to 91.61\%. This advantage is retained
even if we binarize these high-dimensional embeddings via thresholding at 0 into
vectors that have the same memory footprint as 64-dimensional float embeddings.
This further supports the superiority of the softmax-based losses over the
traditional triplet loss. In fact, switching to an EfficientNet-B0 backbone,
a state-of-the-art feature extractor with one fifth of trainable parameters
compared to Resnet50, we achieve superior results even with its smaller
embedding size of 1280. Furthermore, the lack of online sampling reduced
our training times from 8 days down to 33 hours on the same machine,
nearly a 6-fold improvement in computational performance with an
\emph{increase} in the target quality metrics.

\paragraph*{Effect of margin.} Our experiments with a constant margin introduced
on the positive pair (LMCL) already improved the results compared to the vanilla
normalized softmax function. In order to ensure that the witnessed results are
attributed to the margin, we ran several experiments with various temperature
scales and margin values, but found no improvements. This suggests that DML
setups can benefit from large margins combined with temperature scaling similar
to classification problems.

We tried two text-embedding models on the product titles to test the
adaptive-margin method:

\begin{itemize}
	\item average of fastText word embeddings;
	\item sentence embedding with pre-trained AmaBERT.
\end{itemize}

For both cases the performance was better compared to any previous experiment,
with the fastText embeddings reaching 91.64\% and the AmaBERT 92.11\% Recall@1.
Although these results are just marginally higher than the best LMCL setting,
both outperform the baseline by a considerable margin. Again, this advantage
remains present even when the embeddings are further binarized. For more details
please refer to Table~\ref{tab:amazonfashion_amazonfashion}.

\begin{table}[]
	\begin{tabular}{l|lllllll}
		Recall@K & 1 & 5 & 10 & 20 & 30 & 40 & 50  \\ \hline 
		$\text{Triplet loss}^{128}$+ATL+OANNS\cite{Zhao2019AWS}  & 88.46 & 94.85 & 96.27 & 97.41 & 97.88 & 98.22 & 98.40 \Tstrut \\ \hline
		$\text{NormSoftmax}^{2048}\dagger$ & 91.61 & 96.79 & 97.93 & \textbf{98.63} & \textbf{98.92} & \textbf{99.08} & \textbf{99.21} \Tstrut \\
		$\text{NormSoftmax}^{2048bits}\dagger$ & 90.62 & 96.24 & 97.29 & 98.24 & 98.62 & 98.79 & 98.99\\
		$\text{NormSoftmax}^{1280}\ddagger$ & 90.78 & 96.47 & 97.56 & 98.35 & 98.67 & 98.88 & 99.05  \\
		$\text{NormSoftmax}^{1280bits}\ddagger$ & 90.12 & 95.88 & 97.19 & 98.09 & 98.46 & 98.69 & 98.88 \\
		$\text{NormSoftmax}^{1280}\ddagger \sigma=30$ & 89.46 & 95.70 & 96.88 & 97.77 & 98.11 & 98.40 & 98.62  \\
		$\text{NormSoftmax}^{1280bits}\ddagger \sigma=30$ & 88.61 & 94.99 & 96.27 & 97.32 & 97.81 & 98.10 & 98.34  \\
		$\text{NormSoftmax}^{1280}\ddagger \text{LMCL}$ & 91.89 & 96.53 & 97.57 & 98.26 & 98.53 & 98.70 & 98.85  \\
		$\text{NormSoftmax}^{1280bits}\ddagger \text{LMCL}$ & 91.00 & 96.19 & 97.21 & 97.96 & 98.33 & 98.52 & 98.60 \\
		$\text{NormSoftmax}^{1280}\ddagger \text{fastText}$ & 91.77 & 96.71 & 97.65 & 98.31 & 98.60 & 98.75 & 98.90  \\
		$\text{NormSoftmax}^{1280bits}\ddagger \text{fastText}$ & 91.11 & 96.31 & 97.38 & 98.06 & 98.39 & 98.55 & 98.69  \\
		$\text{NormSoftmax}^{1280}\ddagger \text{AmaBERT}$ & 91.64 & 96.70 & 97.66 & 98.31 & 98.57 & 98.79 & 98.96  \\
		$\text{NormSoftmax}^{1280bits}\ddagger \text{AmaBERT}$ & 90.85 & 96.29 & 97.39 & 98.05 & 98.37 & 98.58 & 98.74  \\
		$\text{NormSoftmax}^{2048}\dagger \text{AmaBERT}$ & \textbf{92.11} & \textbf{96.76} & \textbf{97.70} & 98.27 & 98.56 & 98.77 & 98.92  \\
		$\text{NormSoftmax}^{2048bits}\dagger \text{AmaBERT}$ & 91.87 & 96.60 & 97.50 & 98.15 & 98.49 & 98.70 & 98.82 
	\end{tabular}
	\caption{Retrieval performance on Amazon Fashion Retrieval Dataset. $\dagger$: ResNet50, $\ddagger$: EfficientNet-B0 feature extractors.}
	\label{tab:amazonfashion_amazonfashion}
\end{table}

In order to test the generalization capability of our models across different
datasets, we also performed evaluations on the DeepFashion retrieval set.
Even though the adaptive margin with AmaBERT embeddings do perform slightly
better than the baseline (from 77\% to 78.08\% Recall@1), all other
configurations perform significantly worse. This suggests that the softmax-based
approaches somewhat overfit on the training domain, which makes them perform
worse on other datasets.
Detailed results are presented in Table~\ref{tab:amazonfashion_deepfashion}.

\begin{table}[]
	\begin{tabular}{l|llllll}
		Recall@K & 1 & 10 & 20 & 30 & 40 & 50 \\ \hline 
		$\text{Triplet loss}^{128}$+ATL+OANNS\cite{Zhao2019AWS} & 77 & 93.2 & 95.3 & 96.2 & N/A & 97.2 \Tstrut  \\ \hline
		$\text{NormSoftmax}^{2048}\dagger$ & 74.98 & 91.93 & 94.35 & 95.48 & 96.16 & 96.55\Tstrut \\
		$\text{NormSoftmax}^{2048bits}\dagger$ & 72.80 & 90.97 & 93.47 & 94.8 & 95.44 &  96.00 \\
		$\text{NormSoftmax}^{1280}\ddagger$ &  71.12 & 90.95 & 93.92 & 95.16 & 95.98 & 96.49 \\
		$\text{NormSoftmax}^{1280bits}\ddagger$ & 68.76 & 89.53 & 92.87 & 94.32 & 95.16 & 95.84 \\
		$\text{NormSoftmax}^{1280}\ddagger \sigma=30$ & 61.23 & 86.17 & 90.53 & 92.47 & 93.63 & 94.37 \\
		$\text{NormSoftmax}^{1280bits}\ddagger \sigma=30$ & 57.74 & 83.76 & 88.51 & 90.77 & 92.19 & 93.14 \\
		$\text{NormSoftmax}^{1280}\ddagger \text{LMCL}$ & 75.21 & 92.97 & 94.99 & 96.05 & 96.76 & 97.19 \\
		$\text{NormSoftmax}^{1280bits}\ddagger \text{LMCL}$ & 72.57 & 91.36 & 94.06 & 95.28 & 96.01 & 96.53 \\
		$\text{NormSoftmax}^{1280}\ddagger \text{fastText}$ & 75.85 & \textbf{93.30} & \textbf{95.45} & \textbf{96.38} & \textbf{96.98} & \textbf{97.48} \\
		$\text{NormSoftmax}^{1280bits}\ddagger \text{fastText}$ & 73.67 & 91.90 & 94.62 & 95.79 & 96.45 & 96.96 \\
		$\text{NormSoftmax}^{1280}\ddagger \text{AmaBERT}$ & 76.23 & 93.21 & 95.32 & 96.32 & 96.88 & 97.29 \\
		$\text{NormSoftmax}^{1280bits}\ddagger \text{AmaBERT}$ & 74.27 & 92.12 & 94.63 & 95.74 & 96.38 & 96.78 \\
		$\text{NormSoftmax}^{2048}\dagger \text{AmaBERT}$ & \textbf{78.08} & 93.25 & 95.22 & 96.12 & 96.71 & 97.14 \\
		$\text{NormSoftmax}^{2048bits}\dagger \text{AmaBERT}$ & 76.42 & 92.48 & 94.58 & 95.65 & 96.34 & 96.76
	\end{tabular}
	\caption{Retrieval performance on DeepFashion Dataset when trained on the Amazon Fashion Retrieval Dataset. $\dagger$: ResNet50, $\ddagger$: EfficientNet-B0.}
	\label{tab:amazonfashion_deepfashion}
\end{table}

\paragraph*{DeepFashion dataset.}
We trained the best performing models on the DeepFashion dataset to be able to
gauge the performance of the proposed approach against the state of the art.
This dataset however does not contain titles for the classes. Thus, with the
fastText model, we computed the average of the word embeddings of all attributes
per class. With AmaBERT we embedded the first bullet-point description.
Our results show state of the art performance for both models compared to other
recent DML approaches, even after binarizing the resulting embeddings.
The difference between the two text embedders is however very small
(91.79\% vs. 91.9\% Recall@1). We summarized the results on this dataset in
Table~\ref{tab:deepfashion}.

\begin{table}[]
	\begin{tabular}{lllllll}
		Recall@K & 1 & 10 & 20 & 30 & 40 & 50 \\ \hline 
		$\text{Triplet loss}^{128}$+ATL+OANNS\cite{Zhao2019AWS} & 77.03 & 93.22 & 95.34 & 96.24 & 96.87 & 97.25 \Tstrut  \\ \hline
		$\text{NormSoftmax}^{2048}\dagger$ & 89.26 & 97.74 & 98.52 & 98.81 & 98.99 & 99.13 \Tstrut \\
		$\text{NormSoftmax}^{2048bits}\dagger$ & 88.83 & 97.73 & 98.53 & 98.81 & 98.94 & 99.08 \\
		$\text{NormSoftmax}^{2048}\dagger \text{fastText}$ & 91.79 & 98.32 & 98.89 & \textbf{99.16} & \textbf{99.27} & \textbf{99.35} \\
		$\text{NormSoftmax}^{2048bits}\dagger \text{fastText}$ & 90.97 & 98.07 & 98.73 & 98.99 & 99.12 & 99.24 \\
		$\text{NormSoftmax}^{2048}\dagger \text{AmaBERT}$ & \textbf{91.90} & \textbf{98.37} & \textbf{98.94} & 99.13 & 99.22 & 99.28 \\
		$\text{NormSoftmax}^{2048bits}\dagger \text{AmaBERT}$ & 91.43 & 98.28 & 98.84 & 99.00 & 99.13 & 99.23 \\ \hline
		$\text{A-BIER}^{512}$~\cite{BIER} & 83.1 & 95.1 & 96.9 & 97.5 & 97.8 & 98.0 \Tstrut\\
		$\text{ABE}^{512}$~\cite{WonsikAttentionEnsemble} & 87.3 & 96.7 & 97.9 & 98.2 & 98.5 & 98.7 \\
		$\text{Multi-similarity}^{128}$~\cite{wang2019multi} & 88.0 & 97.2 & 98.1 & 98.5 & 98.7 & 98.8 \\
		$\text{Multi-similarity}^{512}$~\cite{wang2019multi} & 89.7 & 98.9 & 98.5 & 98.8 & 99.1 & 99.2
	\end{tabular}
	\caption{Retrieval performance on the DeepFashion Dataset and comparison to state of the art.
            	$\dagger$: ResNet50, $\ddagger$: EfficientNet-B0.}
	\label{tab:deepfashion}
\end{table}

\section{Conclusions}
We have shown how it is possible to introduce adaptive additive margins to a
classification-based loss popular in deep metric learning. We leverage that to
take advantage of additional available data in different modalities and show
how to incorporate text from product titles and attributes during training
using different sentence embedding methods like fastText and BERT. Moreover,
we has demonstrated that this adaptive extension to the classification-loss is
compatible with the use of proxies and that it not only inherits the
computational and simplicity advantages of this combination but pushes it
further, in that it allows us to set a new state-of-the-art for DML-based image
retrieval in both the public DeepFashion In-Shop Clothes Retrieval benchmark
and a larger Amazon-internal Fashion dataset. Our results are consistent across
different image-feature extraction backbones and text embedding models, and
still show improvements when large-dimensional feature vectors are binarized
(allowing sparse and compact feature vectors for indexing).


\section{Acknowledgments}

The authors would like to thank Xiaonan Zhao for sharing the code,
hyperparameters, and the Amazon dataset used for the paper that provided the
basis of this work. We also would like to thank Sergey Sokolov for sharing his
AmaBERT code and model which enabled us to experiment with state-of-the-art
domain-specific text representations.

{\small
	\bibliographystyle{abbrvnat}
	\bibliography{amlc_adaptive_softmax_2020}
}

\end{document}